\newif\ifdraft\drafttrue

\documentclass{article} % For LaTeX2e
\usepackage{iclr2027_conference,times}
\usepackage{wrapfig}

\usepackage{amsmath,amsfonts,bm}

\def\eqref#1{equation~\ref{#1}}
\def\1{\bm{1}}

\DeclareMathAlphabet{\mathsfit}{\encodingdefault}{\sfdefault}{m}{sl}
\SetMathAlphabet{\mathsfit}{bold}{\encodingdefault}{\sfdefault}{bx}{n}

\usepackage{hyperref}
\usepackage{url}
\usepackage{graphicx}
\usepackage{etoc}
\usepackage[utf8]{inputenc} % allow utf-8 input
\usepackage[T1]{fontenc}    % use 8-bit T1 fonts
\usepackage{booktabs}       % professional-quality tables
\usepackage{amsfonts}       % blackboard math symbols
\usepackage{nicefrac}       % compact symbols for 1/2, etc.
\usepackage{microtype}      % microtypography
\usepackage[dvipsnames]{xcolor}         % colors
\usepackage{amsmath}
\usepackage{tcolorbox}
\usepackage{enumitem}
\usepackage{cleveref}
\usepackage{xcolor}

\definecolor{ink}{HTML}{1F1426}
\definecolor{deepemerald}{HTML}{008F68}
\definecolor{softgreen}{HTML}{E8F6F1}
\definecolor{creamgold}{HTML}{A97818}

\hypersetup{
    colorlinks=true,
    linkcolor=deepemerald,
    citecolor=creamgold,
    urlcolor=deepemerald
}

\ifdraft
    \newcommand{\sami}[1]{{\color{blue}[\textbf{Sami:} #1]}}
    \newcommand{\johan}[1]{{\color{cyan}[\textbf{Johan:} #1]}}
\else
    \newcommand{\sami}[1]{}
    \newcommand{\johan}[1]{}
\fi

\iclrpreprint

\title{Freeze, Don't Clone: Fine-Tuning with Expressive Policies Without Retaining Offline Data}

\author{%
\makebox[\textwidth][c]{%
Sami Nur Islam\textsuperscript{*1,2}\quad
Johan Obando-Ceron\textsuperscript{*1,3}\quad
Doina Precup\textsuperscript{1,2,4}
}\\[0.1em]
\makebox[\textwidth][c]{%
\textbf{Aaron Courville}\textsuperscript{1,3,4}\quad
\textbf{Pablo Samuel Castro}\textsuperscript{1,3}
}\\[1em]
\makebox[\textwidth][c]{%
\textsuperscript{*}Equal contribution\quad
\textsuperscript{1}Mila -- Qu\'ebec AI Institute\quad
\textsuperscript{2}McGill University\quad
}\\
\makebox[\textwidth][c]{%
\textsuperscript{3}Universit\'e de Montr\'eal
\textsuperscript{4}CIFAR AI Chair\quad
}\\[1em]
\makebox[\textwidth][c]{%
\texttt{mohammad.sami.islam@mail.mcgill.ca}
\texttt{jobando0730@gmail.com}
}\\
\makebox[\textwidth][c]{%
\texttt{\{precupdo, courvila, pablo-samuel.castro\}@mila.quebec}
}
}

\makeatletter
\def\addcontentsline#1#2#3{%
\addtocontents{#1}{\protect\contentsline{#2}{#3}{\thepage}{\@currentHref}{}}}
\makeatother

\begin{document}

\etocdepthtag.toc{mtoc}
\etocsettagdepth{mtoc}{none}
\etocsettagdepth{appendix}{none}

% 2. Generate the invisible TOC (Only the print command goes in the box)
\newbox\dummybox
\setbox\dummybox\vbox{%
  \tableofcontents 
}

\maketitle

\begin{abstract}
Offline-to-online reinforcement learning uses offline pretraining to
accelerate subsequent online learning. While many methods assume the
offline dataset remains available throughout fine-tuning, there are many settings where this is not the case. Prior
work attributes failures in this setting primarily to critic recalibration
under distribution shift. We identify a complementary actor-side failure:
once the offline data disappears, continued behavioral cloning (BC) no longer
preserves the offline behavior distribution and instead refits the behavioral
model to the agent's limited, and nonstationary, online experience. We demonstrate that stopping online behavioral fitting while preserving the
pretrained expressive prior substantially improves no-retention fine-tuning across implicit
flow Q-learning (IFQL), flow Q-learning (FQL), and action-chunked RL (QC) on seven
OGBench domains. We trace this failure to a behavioral feedback loop: continuing online behavioral cloning rapidly erases the pretrained prior, inducing representational collapse and action diversity loss that starves the critic of useful exploration. Freezing the prior breaks this cycle, preserving offline coverage while allowing value learning to drive continuous improvement. \textbf{Code is available in
\href{https://anonymous.4open.science/r/bc-free-finetuning-3BB4/}{BC-Free Fine-Tuning}.}

\end{abstract}

\section{Introduction}
\label{sec:introduction}
Offline pretraining should allow an agent to learn from a dataset once and
then improve through interaction without carrying that dataset forever. Yet
most offline-to-online RL methods continue sampling offline transitions during
online fine-tuning. This assumption can be expensive or infeasible when the
original data are constrained by storage, privacy, licensing, or deployment
requirements. Worse, simply deleting the offline buffer can cause an agent to
rapidly lose the capabilities acquired during pretraining. We ask how an
offline-pretrained agent should be fine-tuned when the original dataset is no
longer available, which hereafter we refer to as ``no-retention fine-tuning''.

Previous work identifies the critic as an important source of instability in
this setting. The transition to online interaction abruptly changes the
state--action distribution, forcing value estimates learned from broad offline
data to be recalibrated from a small online replay buffer. WSRL stabilizes this
transition by collecting initial rollouts with the pretrained policy before
beginning online updates \citep{zhou2025efficient}. This critic-centered view
explains an important part of no-retention fine-tuning, but it leaves open what
should happen to the behavioral model itself.

\begin{figure}[t]
    \centering
    \includegraphics[width=0.63\textwidth]{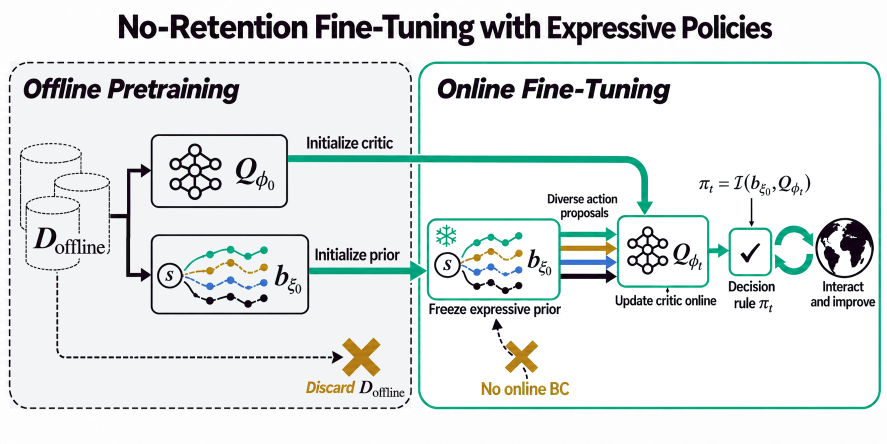}%
    \includegraphics[width=0.37\linewidth]{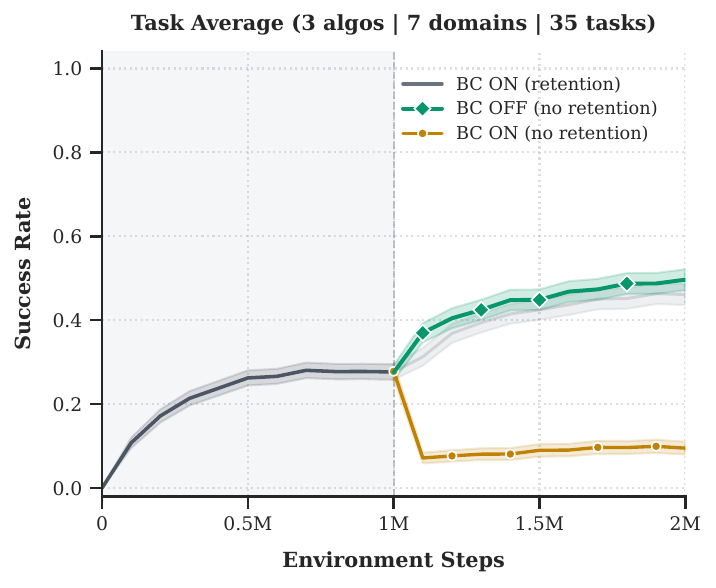}
    \vspace{-0.8cm}
    \caption{
    \textbf{Stopping BC enables no-retention fine-tuning.}
    \textbf{Left:} After offline pretraining, we freeze the behavioral prior and
    continue updating the critic using online data.
    \textbf{Right:} Aggregate performance across seven OGBench tasks after $1$M
    offline gradient steps (gray) and $1$M online environment steps (white).
    Stopping BC consistently improves methods with expressive policies.  }
    \label{fig:main_results}
    \vspace{-0.2cm}
\end{figure}

Behavioral regularization is central to many offline RL methods. By fitting or
remaining close to dataset actions, the policy is discouraged from exploiting
values assigned to unsupported actions
\citep{kumar2019stabilizing,wu2019behavior,fujimoto2021minimalist,tarasov2023revisiting}.
Recent approaches combine this principle with diffusion or flow-based
policies that can represent complex and multimodal behavior distributions
\citep{hansenestruch2023idql,park2025flow,li2025actionchunking}. While the
offline data remain available, continuing behavioral fitting can keep the
policy anchored to the distribution learned during pretraining. When the offline dataset disappears, however, this regularization can become a liability. Behavioral cloning no longer preserves the offline
distribution; it trains the behavioral model to imitate actions generated by
the agent's current decision rule. Early online experience is scarce,
correlated, policy-dependent, and nonstationary. Consequently, behaviors that
are not immediately revisited may be undervalued and underused
before improved value estimates can (re)highlight their value. This motivates a simple
principle: \emph{keep learning which actions are {\em valuable}, but stop relearning
which actions are {\em possible}}.

We instantiate this principle by stopping online behavior cloning (BC) to preserve the expressive prior learned offline. This does not freeze the
entire agent: the critic and algorithm-specific improvement mechanisms continue
learning from online experiences. IFQL makes this separation particularly clear:
during the online stage a fixed flow policy proposes actions, while an updated critic improves which proposal
is selected. In FQL, the flow policy remains fixed while the critic and a
one-step actor continue learning, with the one step actor distilled toward the frozen
flow policy. Action-chunked RL (QC) similarly preserves a prior over temporally extended
behaviors while updating their values and selections.

Across seven OGBench domains (35 tasks), preserving the pretrained prior substantially improves no-retention fine-tuning for IFQL, FQL, and QC. However, our ablations reveal that stopping BC alone is insufficient without an expressive architecture: the frozen prior must support complex, multimodal distributions to succeed. When the prior is restricted to a simple Gaussian, performance collapses. These results complement critic-centered accounts with a clear policy-side design principle: freeze an expressive, multimodal behavioral prior, and ensure online value learning can directly exploit it.

\section{No-Retention Fine-Tuning with Expressive Priors}
\label{sec:method}

\paragraph{Problem Setting}
\label{sec:problem_setting}

We consider a discounted Markov decision process
$\mathcal M=(\mathcal S,\mathcal A,P,r,\gamma)$ and seek a policy maximizing
$J(\pi)=\mathbb E_{\pi,P}[\sum_{t=0}^{\infty}\gamma^t r(s_t,a_t)]$.
Offline pretraining uses
$\mathcal D_{\mathrm{off}}=\{(s_i,a_i,r_i,s_i')\}_{i=1}^{N}$ to learn a
critic $Q_{\phi_0}$ and a policy or behavioral model. During online
fine-tuning, the agent accumulates
$\mathcal D_{\mathrm{on}}^t$ from interaction. We study the no-retention
setting, in which $\mathcal D_{\mathrm{off}}$ is unavailable after
pretraining and all online updates use only $\mathcal D_{\mathrm{on}}^t$. Let $d_{\mathrm{off}}$ and $d_{\mathrm{on}}^t$ denote the empirical
state--action distributions of the two buffers. Retained-data fine-tuning may
train on $d_t^\rho=\rho_t d_{\mathrm{off}}+(1-\rho_t)d_{\mathrm{on}}^t$ for
$\rho_t>0$. No retention sets $\rho_t=0$. This change matters because a
behavioral loss optimized under $d_t^\rho$ remains partly supervised by
offline behavior, whereas the same loss under $d_{\mathrm{on}}^t$ fits only
the agent's evolving online experience.

\paragraph{Behavioral Priors and Decision Policies.}
\label{sec:behavioral_priors}

We distinguish a \emph{behavioral prior} $b_\xi(a\mid s)$ from the
\emph{decision policy} used for interaction. The prior models the conditional
action distribution represented in its training data. For a likelihood-based
model, behavioral fitting can be written as
$\mathcal L_{\mathrm{BC}}(\xi;\mathcal D)=-\mathbb E_{(s,a)\sim\mathcal D}
[\log b_\xi(a\mid s)]$; deterministic, diffusion, and flow policies use the
corresponding regression or generative objectives. The decision policy need not equal $b_\xi$. It may be represented by a
separate value-optimized actor or constructed by critic-guided selection from
the prior. We write this relationship as
$\pi_t=\mathcal{I}_{\eta_t}(b_{\xi_t},Q_{\phi_t})$, where $\mathcal{I}$ is a
policy-improvement operator and $\eta_t$ denotes any additional trainable
parameters. Thus, fixing the prior need not fix the decision rule, which can still adapt through $Q_{\phi_t}$ or $\eta_t$.

\paragraph{Flow-Based Behavioral Priors}
\label{sec:flow_priors}

A diagonal Gaussian is efficient but unimodal and may poorly model datasets
containing several plausible actions at the same state. Flow policies instead
transport noise $z\sim\mathcal N(0,I)$ to the behavior distribution using a
state-conditioned velocity field
\citep{liu2022flow,lipman2023flow,park2025flow}. For a dataset action $a$,
flow time $u\sim\mathcal U[0,1]$, and interpolation $x_u=(1-u)z+ua$, conditional flow matching minimizes
$\mathcal{L}_{\mathrm{FM}}(\xi;\mathcal{D})
=\mathbb{E}_{(s,a),z,u}
[\|f_\xi(s,x_u,u)-(a-z)\|_2^2]$.
Integrating $\mathrm dx_u/\mathrm du=f_\xi(s,x_u,u)$ from $x_0=z$ to
$u=1$ produces $F_\xi(s,z)$, whose induced distribution defines
$b_\xi(\cdot\mid s)$. Flow matching therefore plays the role of BC: it fits
the behavioral prior to actions represented in the current replay data. IFQL uses this flow directly as a proposal distribution. In the best-of-$K$ variant studied here, IFQL samples
$a^{(i)}\sim b_\xi(\cdot\mid s)$ and executes
$a^\star=\arg\max_i Q_\phi(s,a^{(i)})$. FQL instead retains $F_\xi$ as a
behavioral reference and trains a one-step actor $\mu_\psi(s,z)$ by minimizing
$\mathcal L_{\mathrm{FQL}}(\psi)
=\mathbb E_{s,z}[-Q_\phi(s,\mu_\psi(s,z))
+\alpha\|\mu_\psi(s,z)-\operatorname{sg}[F_\xi(s,z)]\|_2^2]$.
Thus, FQL separates flow fitting from value-guided actor distillation.

\paragraph{Preserving the Prior During Online Fine-Tuning}
\label{sec:preserving_prior}

After offline pretraining, we freeze the behavioral prior,
$\xi_t=\xi_0$, while updating the critic and algorithm-specific improvement
parameters on $\mathcal D_{\mathrm{on}}^t$. The decision policy is therefore
$\pi_t=\mathcal I_{\eta_t}(b_{\xi_0},Q_{\phi_t})$. We denote continued
behavioral fitting on online replay by \textsc{BC-On} and freezing the prior by
\textsc{BC-Off}. Stopping BC affects only behavioral-model fitting; value
learning, action selection, and distillation toward the frozen prior remain
active. Under \textsc{BC-Off}, IFQL freezes its flow and updates the critic; FQL
additionally updates its distilled one-step actor; and QC \citep{li2025actionchunking} freezes its
flow-based chunk prior while updating the chunk-level critic used for
best-of-$N$ selection.

\paragraph{Experimental Setup and Evaluation Protocol.}
Our primary benchmark spans seven OGBench manipulation and locomotion domains (\texttt{cube-double-play}, \texttt{puzzle-4x4-play}, \texttt{scene-play}, \texttt{antsoccer-arena=navigate}, \textbf{antmaze-large-navigate}, \texttt{humanoidmaze-medium-navigate}, \texttt{humanoidmaze-large-navigate}), evaluating all five single-task variants per domain. For targeted extensions, we evaluate on low-quality data using \texttt{cube-double-noisy} and \texttt{antmaze-large-explore}, and examine human teleoperation demonstrations using the Multi-Human (\texttt{MH}) \texttt{Can} and \texttt{Lift} tasks from Robomimic \citep{robomimic2021}. Algorithms are pretrained for 1M gradient steps and fine-tuned for 1M environment steps; under no retention, the offline buffer is discarded at the transition. We compare \textsc{BC-On} versus \textsc{BC-Off} across IFQL, FQL, and QC, focusing detailed diagnostics on IFQL and FQL. All evaluations use three independent seeds per task; we report mean success rates with shaded standard error (SE) bands. Full environment specifications, task selection rationale, hyperparameters, and per-task results are detailed in \autoref{app:hyper}.

\section{Freeze the Prior, Improve the Decision Rule}
\label{sec:main_results}

We first show that freezing the behavioral prior improves no-retention
fine-tuning with flow-based policies. We then test whether removing BC is
sufficient without an expressive prior and whether freezing remains effective
after pretraining on low-quality data.

While our primary evaluations include QC, we focus our detailed diagnostic ablations on IFQL and FQL. In addition to their strong performance across recent offline-to-online RL literature \citep{park2025flow, li2026qlearningadjointmatching}, these two algorithms cleanly span the two fundamental mechanisms  for utilizing an expressive prior: direct critic-guided action selection (IFQL) and actor distillation (FQL), while avoiding confounding factors from temporal abstraction.

\paragraph{Stopping BC improves no-retention fine-tuning.}

\begin{figure}[t]
    \centering
    \includegraphics[width=\textwidth]{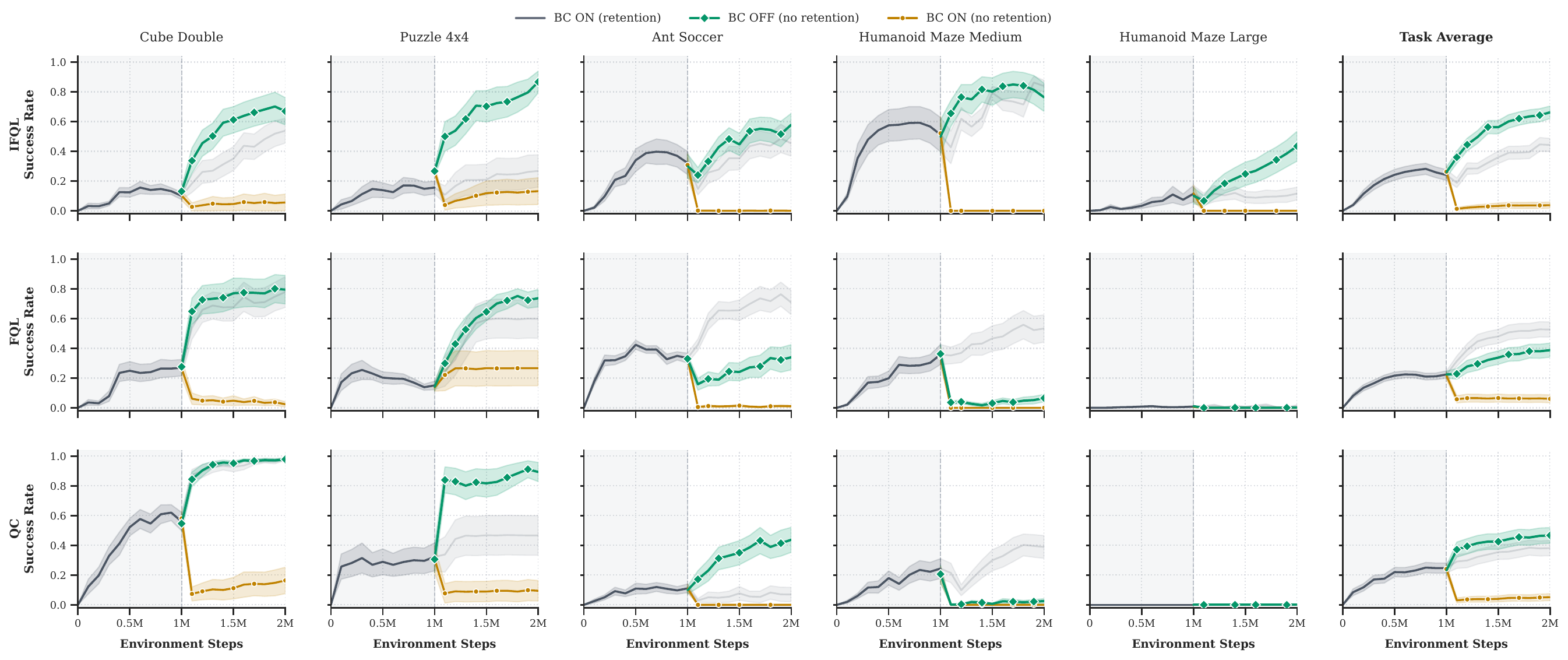}
    \vspace{-0.8cm}
    \caption{
    \textbf{Stopping BC improves no-retention fine-tuning across expressive policy classes.} 
    Performance of IFQL, FQL, and QC (Action-Chunked RL) across five OGBench domains (25 tasks). The vertical dashed line marks the transition from offline pretraining to online fine-tuning. Continuing BC without offline data (\textsc{BC-On}, orange dashed) causes catastrophic performance collapse. \textsc{BC-Off} (red dashed) rescues online fine-tuning and often approaches the performance of the standard retained-data baseline (blue solid). The full plot with the seven domains can be found in the Appendix~\ref{app:more_domains}.
    }
    \label{fig:bc_main_results}
\end{figure}

\autoref{fig:bc_main_results} compares \textsc{BC-On} and
\textsc{BC-Off} during no-retention fine-tuning. Across IFQL, FQL, and
action-chunked RL, stopping behavioral fitting consistently improves
online learning. The largest gains occur for IFQL:
\textsc{BC-Off} improves the aggregate mean and outperforms \textsc{BC-On} across tasks.
FQL and the action-chunked methods exhibit the same overall trend,
although the magnitude of the improvement varies across algorithms and
domains. The IFQL and QC results are particularly revealing. Neither directly
optimizes their behavioral flow (or a distilled version of it) for return; the flow proposes candidate
actions, and the critic selects among them. Under \textsc{BC-Off}, the
proposal distribution remains fixed while the critic continues learning
from online experience. Performance can therefore improve without
refitting the behavioral model. Under \textsc{BC-On}, the same flow matching loss is
continually optimized on online-only replay, and performance is
substantially lower. These results show that an objective useful during offline pretraining
need not remain useful after its original data disappear. In this
setting, preserving the pretrained behavioral model is more effective
than continually refitting it to the agent's recent experience.

    \begin{wrapfigure}{r}{0.45\linewidth}
        \centering
        \vspace{-0.8cm}
        \includegraphics[width=\linewidth]{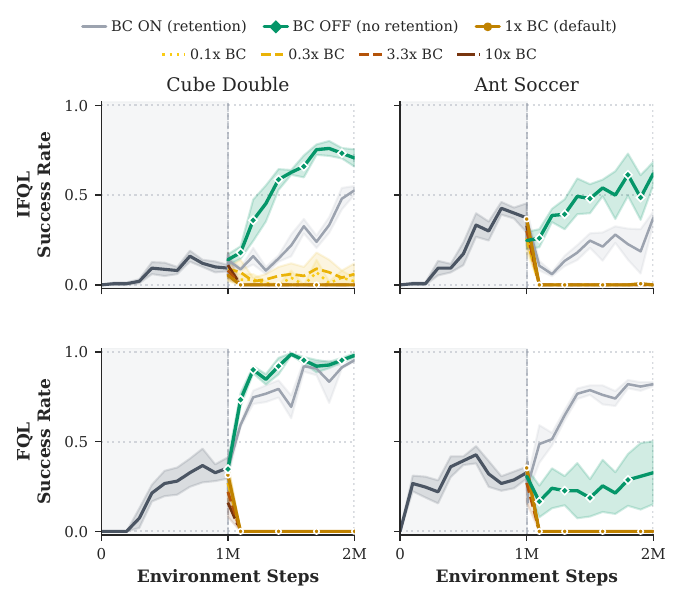}
        \vspace{-0.8cm}
        \caption{
        \textbf{Sensitivity to the BC loss weight.}
        Online fine-tuning performance across varying BC loss weights ($0.1\times$ to $10\times$ default) without offline data retention.
        Scaling the BC weight up or down (yellow to dark brown) fails to prevent collapse in both IFQL and FQL.
        Only removing the BC loss entirely (\textsc{BC-Off}, green) avoids collapse and enables sustained online learning.
        }
        \label{fig:bc_weight_sensitivity}
        \vspace{-0.6cm}
    \end{wrapfigure}

\paragraph{Attenuating the BC penalty cannot prevent collapse.}
To determine whether the failure of online behavioral cloning is merely an artifact of an overly aggressive regularization weight, we evaluate no-retention fine-tuning across varying BC penalty magnitudes ($0.1\times$ to $10\times$ the default). \autoref{fig:bc_weight_sensitivity} demonstrates that scaling the BC weight, whether amplifying or aggressively attenuating it, fails to prevent catastrophic collapse in both IFQL and FQL. This confirms the failure mode is fundamentally tied to the objective itself rather than its hyperparameter tuning: any active optimization toward the agent's narrowing online experience progressively erases the offline prior. Consequently, only the complete removal of the behavioral loss (\textsc{BC-Off}) enables sustained online improvement.

\begin{figure*}[t]
    \centering
    \includegraphics[width=\textwidth]{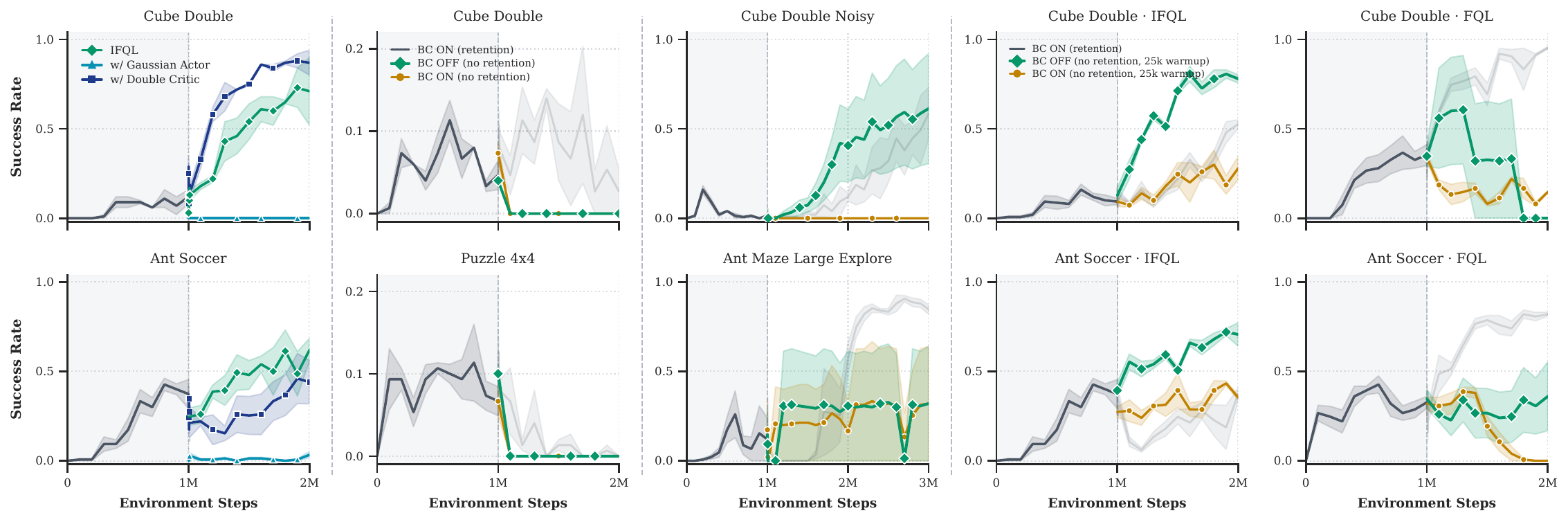}
    \vspace{-0.7cm}
    \caption{
    \textbf{When does freezing the behavioral prior enable no-retention fine-tuning?}
    \textbf{Col 1:} In IFQL, fine-tuning succeeds with a standard Bellman double critic (navy) but collapses entirely when its expressive flow prior is ablated in favor of a Gaussian actor                     (blue).
    \textbf{Col 2:} Removing ReBRAC's BC penalty (\textsc{BC-Off}) does not prevent collapse (scaled $0.0-0.2$ to highlight the immediate drop), demonstrating that omitting BC without a decoupled prior is insufficient.
    \textbf{Col 3:} A frozen flow prior supports continued online improvement without retention even after pretraining on low-quality offline data.
    \textbf{Cols 4 \& 5:} Seeding replay with 25k online transitions prevents \textsc{BC-On} from flatlining across both IFQL and FQL, yet \textsc{BC-Off} consistently achieves superior asymptotic success.
    }
    \label{fig:prior_ablations}
\end{figure*}

\paragraph{Can a large offline warmup rescue continued BC?}
Because the failure of \textsc{BC-On} is directly tied to discarding offline data, we test whether a temporary data ``warmup'' can prevent the collapse. We pre-populate the online buffer using the pretrained checkpoint with 25,000 online transitions. As shown in (\autoref{fig:prior_ablations}, Col 4 \& Col 5), this intervention prevents the catastrophic failure of \textsc{BC-On}; performance no longer flattens at near zero. However, across both IFQL and FQL on \texttt{cube-double} and \texttt{antsoccer-arena}, combining this warmup with \textsc{BC-Off} still yields vastly superior online adaptation. This confirms that continued BC can fundamentally restrict the ceiling of policy improvement even when provided with healthy data initialization.

\paragraph{Dissecting IFQL: Isolating the mechanisms of recovery.}
Our previous experiments reveal that IFQL achieves the most robust performance under the \textsc{BC-Off} regime. To understand whether this success is driven by IFQL's conservative value updates or its generative policy, we isolate these mechanisms by ablating both (\autoref{fig:prior_ablations}, Col 1). We first ablate the critic by replacing IFQL's IQL-style expectile regression with a standard Bellman double Q-learning loss. Despite this modification, the agent still achieves continuous online improvement when offline data is discarded, and BC is removed. Next, we ablate the actor by replacing the expressive flow-matching model with a standard Gaussian policy. Under this configuration, online performance completely flatlines. This component isolation demonstrates that IFQL's resilience is not strictly dependent on its specific value-update mechanism but rather on the expressive capacity of its prior.

\paragraph{Unimodal policies cannot be rescued by removing BC.}
If an expressive prior is truly the bottleneck, then stopping BC should similarly fail to rescue other algorithms restricted to a unimodal representation. To test this, we apply the same no-retention interventions to ReBRAC, an offline RL method utilizing a standard unimodal Gaussian actor. As shown in (\autoref{fig:prior_ablations}, Col 2), discarding the offline dataset causes ReBRAC to completely collapse. Crucially, disabling behavioral cloning (\textsc{BC-Off}) fails to rescue the policy; performance flat lines at zero alongside \textsc{BC-On}. This provides further evidence that successfully fine-tuning without offline data may require not only stopping BC but also a sufficiently expressive prior. Note that to ensure a fair evaluation, the ReBRAC plots feature the Puzzle domain instead of Ant Soccer, as ReBRAC's offline performance on Ant Soccer was near zero.

\begin{wraptable}{r}{0.53\textwidth}
\vspace{-1cm}
\centering
\small
\setlength{\tabcolsep}{4pt}
\renewcommand{\arraystretch}{0.95}
\caption{\textbf{Earlier BC termination performs best.}
Final success after $1$M online steps.}
\label{tab:bc_stopping_times}
\begin{tabular}{lcc}
\toprule
$\tau$
& \shortstack{\texttt{cube-}\\\texttt{double}}
& \shortstack{\texttt{antsoccer-}\\\texttt{arena}} \\
\midrule
$0$ (immediate) & \textbf{0.85} & \textbf{0.71} \\
$200$k          & 0.71 & 0.49 \\
$500$k          & 0.53 & 0.50 \\
$800$k          & 0.61 & 0.51 \\
$1$M (never)    & 0.40 & 0.19 \\
\bottomrule
\end{tabular}
\vspace{-0.3cm}
\end{wraptable}

\paragraph{Dissecting FQL: Distillation is not an optional regularizer.}
\label{app:distillation}

Under BC-OFF, FQL freezes the flow network but continues distilling its one-step actor toward it. If we sever this link by disabling the distillation loss—forcing the actor to train solely on the value objective—performance catastrophically collapses to zero (Table~\ref{tab:fql_true_freeze}). Unlike IFQL, which samples directly from the flow, FQL relies entirely on distillation to transfer knowledge to the deployed policy. Disconnecting this pathway leaves the actor with no access to the offline manifold. This collapse validates our core design principle: preserving a rich behavioral prior drives online success only when the architecture maintains an active pathway for the decision rule to exploit it.

\paragraph{Can \textsc{BC-Off} recover performance from low-quality offline data?}
A potential concern with freezing the behavioral prior is that it permanently anchors the agent to the quality of the offline dataset. We test this by pretraining IFQL on low-quality, sub-optimal datasets (\texttt{cube-double-noisy} and \texttt{antmaze-large-explore}). In these settings, \textsc{BC-Off} still demonstrates a promising capacity for online policy improvement, achieving returns above the initial offline baseline (\autoref{fig:prior_ablations}, Col 3). Because the flow-based policy is highly expressive, it models the broad, noisy distribution of the offline data without collapsing into a single suboptimal mode. Crucially, it appears to capture enough of the underlying valid action manifold to propose a diverse menu of actions that physically obey the environment dynamics, even if those actions originally led to low reward. We hypothesize that during online fine-tuning, this frozen prior continues to constrain the search space to these physically valid candidates. As the critic improves on new online experiences, it learns to consistently identify and select the higher-value actions from this broad distribution, yielding a better decision rule. 

\begin{wrapfigure}{r}{0.48\textwidth}
    \vspace{-0.8cm}
    \centering
    \includegraphics[width=\linewidth]{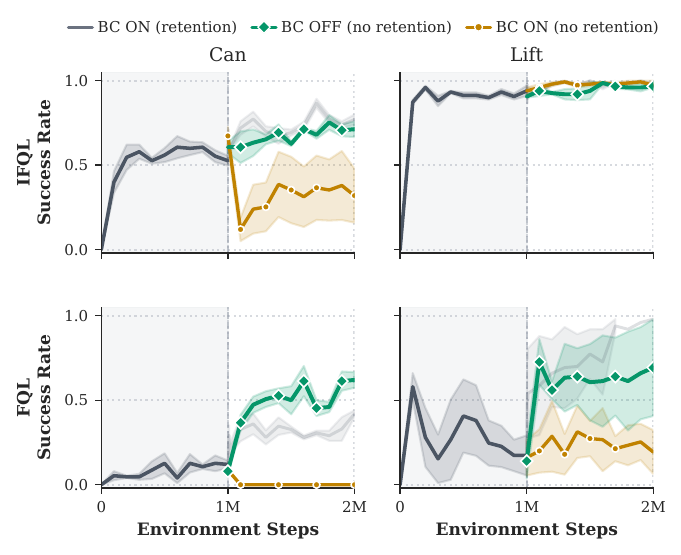}
    \vspace{-0.95cm}
    \caption{\textbf{Online fine-tuning on Robomimic under no data retention.} Removing the BC objective (\textsc{BC-Off}) prevents catastrophic degradation and matches offline retention performance across human demonstrations.}
    \label{fig:robomimic_experiments}
    \vspace{-10pt}
\end{wrapfigure}

\paragraph{Generalization to Human Teleoperation Demonstrations.}
To confirm that prior collapse is not an artifact of synthetic OGBench datasets, we evaluate on the Robomimic benchmark using the Multi-Human (\texttt{MH}) datasets for \texttt{Can} and \texttt{Lift}. These datasets comprise 300 demonstrations collected across six operators of varying proficiency, introducing realistic suboptimality and behavioral multimodality. Details can be found in found in ~\cref{app:robomimic}. As shown in~\autoref{fig:robomimic_experiments}, the same failure mode emerges: continuing online behavioral cloning without offline data retention (\textsc{BC-On}) induces severe degradation and task collapse across both IFQL and FQL. Conversely, discarding the behavioral objective while retaining the expressive prior (\textsc{BC-Off}) completely prevents collapse and allows effective finetuning.

\paragraph{Stopping BC late still recovers part of the freezing benefit.}
Does preserving the behavioral prior require disabling BC from the beginning of online fine-tuning, or can the intervention remain effective after the prior has already been updated on online data? We evaluate intermediate behavioral cloning (BC) termination steps $\tau \in \{200\text{k}, 500\text{k}, 800\text{k}\}$ during $1\text{M}$ online fine-tuning steps to determine if preserving the behavioral prior requires immediate freezing ($\tau=0$). Table~\ref{tab:bc_stopping_times} demonstrates that immediate freezing consistently maximizes performance, whereas persistent BC ($\tau=1\text{M}$) yields the lowest success rates. Terminating BC at any intermediate $\tau$ recovers a substantial fraction of this gap. This demonstrates that removing the behavioral objective can act as a robust stabilizing intervention, even after the prior has begun to degrade.

\begin{wrapfigure}{r}{0.48\linewidth}
    \vspace{-2.9em}
    \centering
    \includegraphics[width=\linewidth]{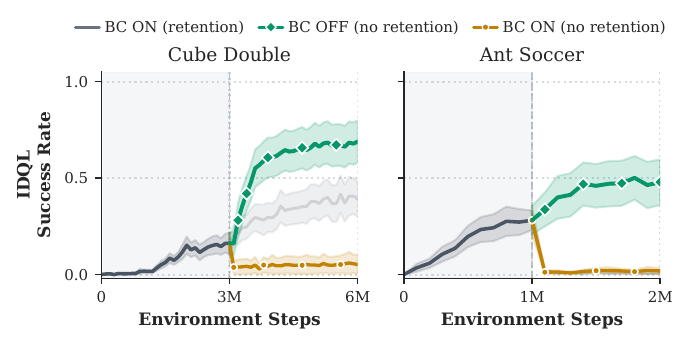}
    \vspace{-1.8em}
    \caption{\textbf{Effect of turning BC off with diffusion priors (IDQL).} Performance across \texttt{Cube Double} and \texttt{Ant Soccer}.}
    \label{fig:idql_ablation}
    \vspace{-1.2em}
\end{wrapfigure}

\paragraph{Generality Across Generative Priors.}
Having shown that turning BC off enables online fine-tuning in flow-matching algorithms in the no offline data retention regime, we examine whether this strategy
extends to other expressive generative priors. To test this, we evaluate Implicit Diffusion Q-Learning (IDQL) \citep{hansen2023idql}, which pairs the implicit RL
framework of IFQL with a diffusion objective (DDPM) rather than flow matching to parameterize its behavioral prior.

As shown in Figure~\ref{fig:idql_ablation}, turning BC off enables consistent online adaptation for diffusion-based policies. In \texttt{Cube Double}, freezing the
pretrained diffusion prior allows the policy to improve substantially beyond the offline checkpoint level, outperforming the standard buffer-retention baseline. In
\texttt{Ant Soccer}, \textsc{BC-Off} yields steady online progress, whereas continuing to train the prior even with full buffer retention degrades performance toward
zero. These results indicate that the benefit of turning BC off is not specific to flow matching, and that a frozen pretrained prior supports effective online fine-tuning across distinct generative modeling paradigms.
  
\section{Diagnosing Online Collapse}
\label{sec:analysis}

The previous section demonstrates that stopping BC can be important for successful no-retention fine-tuning for expressive flow-based policies. But why does continued behavioral cloning fail so catastrophically when offline data is discarded? In this section, we investigate the underlying mechanics of this phenomenon. We trace the failure of BC-ON to a specific behavioral causal chain, moving from catastrophic forgetting of the prior to representational collapse and ultimately critic degeneration. Beyond this feedback loop, we map the degradation of the actor network by tracking progressive layer wise parameter drift. We conclude by ruling out stale optimization statistics to confirm the failure is strictly objective-driven.

\begin{figure}[t]
    \centering

    % Shared legend
    \includegraphics[width=0.55\linewidth]{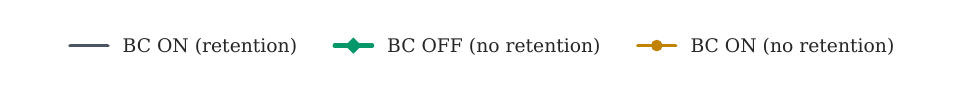}

    \vspace{0.1em}

    % Three panels in one row
    \includegraphics[width=0.24\linewidth]{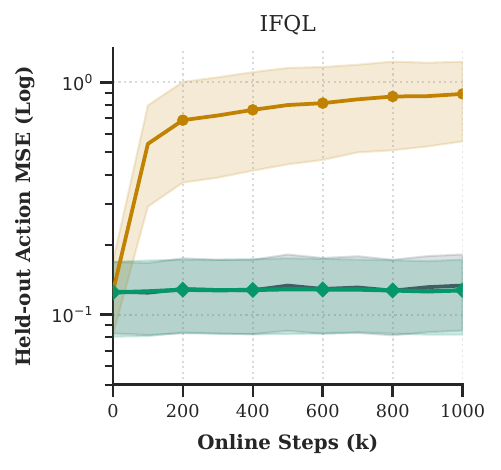}
    \hfill
    \includegraphics[width=0.24\linewidth]{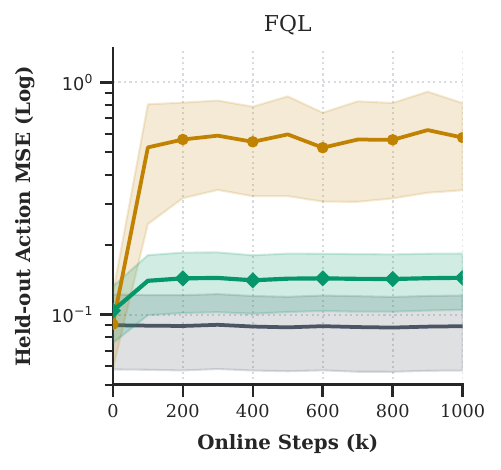}
    \hfill
    \includegraphics[width=0.5\linewidth]{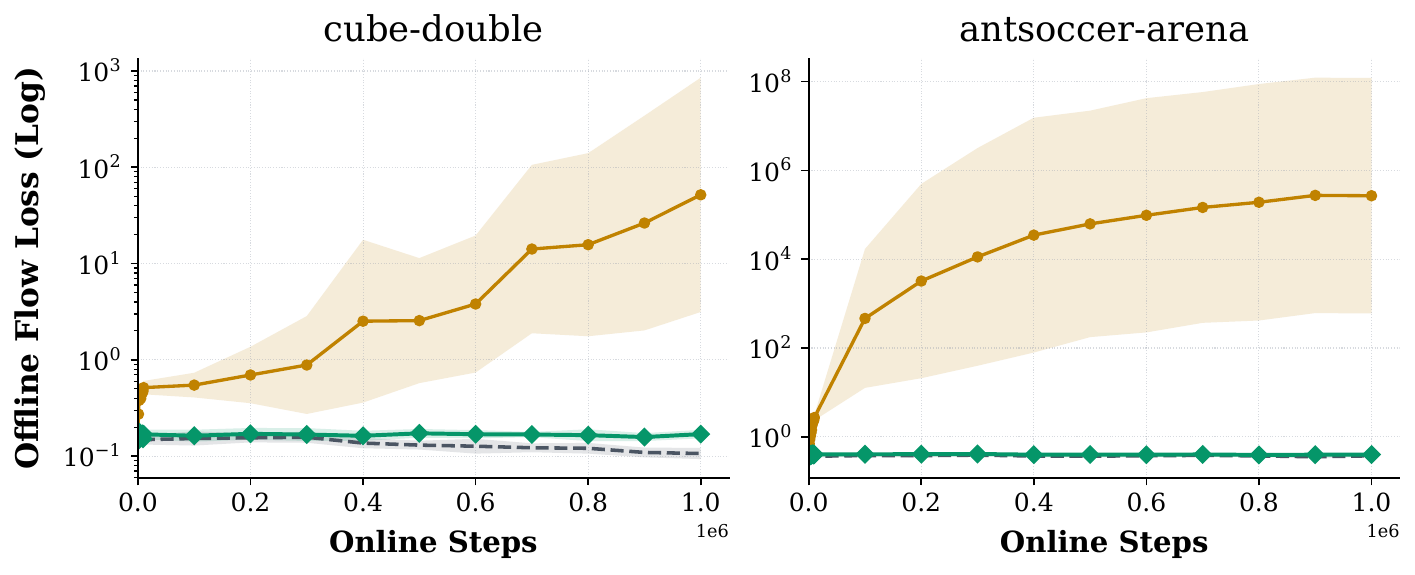}
    \vspace{-0.5cm}
    \caption{
    \textbf{Continued online-only BC forgets the pretrained behavior.}
    \textbf{Left and middle:} Held-out offline action MSE for IFQL and FQL, respectively. Continuing BC without retaining offline data sharply increases error, whereas \textsc{BC-Off} preserves the pretrained behavior.
    \textbf{Right:} BC Loss on the offline data (log scale).
    }
    \label{fig:forgetting_prior}
\end{figure}
\paragraph{Catastrophic forgetting of the behavioral prior.}

When offline data is discarded, the online replay buffer initially contains only a narrow, correlated slice of recent experience. Continuing to optimize the BC loss on this restricted dataset forces the policy to rapidly overfit to its own early, suboptimal rollouts. This induces catastrophic forgetting of the broad, diverse behavior learned during pretraining. We can directly observe this erasure by tracking the flow policy's action Mean Squared Error (MSE) on a held-out offline dataset (\autoref{fig:forgetting_prior}, Left). For both IFQL and FQL, continuing BC without retention causes the offline action MSE to spike immediately and dramatically within the first 100k online steps. The actor rapidly unlearns the valid task manifold. In contrast, removing the behavioral objective (\textsc{BC-Off}) completely preserves this foundational knowledge. This is also captured by the BC loss on the offline data (\autoref{fig:forgetting_prior}, Right).

\begin{figure}[t]
    \centering
    \vspace{-0,5cm}
    \includegraphics[width=\linewidth, height=0.25\textheight, keepaspectratio]{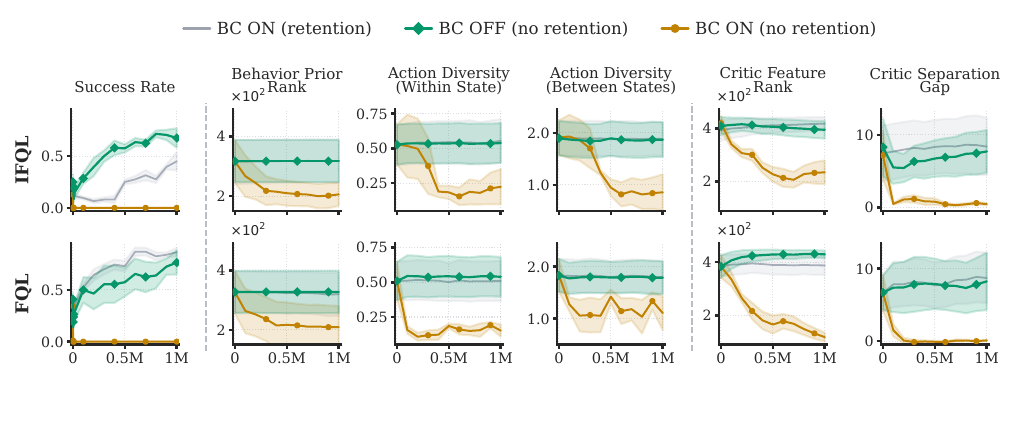}
    \vspace{-1.2cm}
    \caption{
    \textbf{Online BC induces actor collapse and critic data starvation.} 
    \textit(Middle) When BC remains active without offline data, the behavior prior's internal representations degenerate (feature rank collapses), which destroys the policy's cross-state and within-state diversity across both IFQL and FQL. 
    \textit{(Right)} Confined to narrow data collection, the critic is starved of diverse, high-value trajectories. This causes the critic's own feature rank to collapse, destroying its ability to accurately separate high-return states from low-return states (shrinking quantile gap). \textsc{BC-Off} rescues this entire chain. The details of how these metrics are computed are covered in Appendix \ref{app:metrics_definitions}.
    }
    \label{fig:dynamics_actor_critic}
\end{figure}

\begin{figure}[t]
    \centering
    % Left panel: State coverage
    \includegraphics[width=0.65\linewidth]{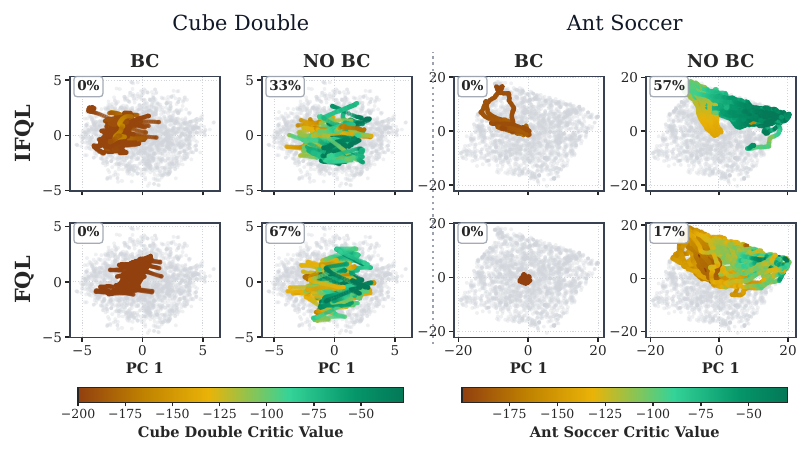}
    \hfill
    % Right panel: Temporal coherency (displacement)
    \raisebox{0.0cm}{\includegraphics[width=0.33\linewidth]{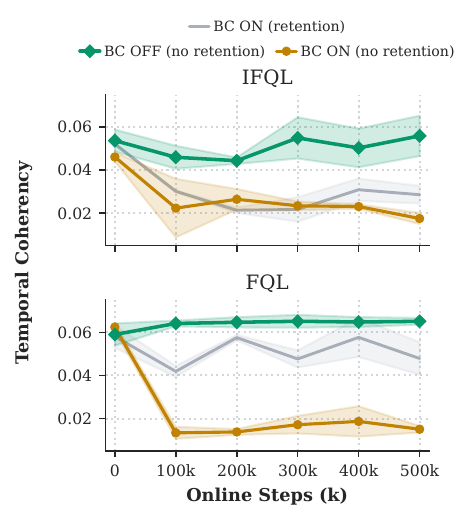}}
    
    \vspace{-0.5cm}
    \caption{
    \textbf{Spatial collapse and degraded exploration under online Behavioral Cloning.} 
    \textbf{Left:} PCA projections of online rollouts onto the offline manifold at 1.2M steps. Stripped of its exploratory capacity by BC, the agent is forced into repetitive failure modes, leading to severely restricted state-space coverage. 
    \textbf{Right:} Short-horizon state displacement on \texttt{cube-double}. Continued BC restricts the agent to increasingly localized behavior, while \textsc{BC-Off} preserves purposeful, wide-area navigation.
    }
    \label{fig:state_coverage}
    \vspace{-0.3cm}
\end{figure}

\paragraph{Online BC induces an actor collapse that drives data drift.}
This catastrophic forgetting of the prior thoroughly destabilizes the entire offline-to-online transition. To understand the structural impact of this failure, we isolate the actor’s intrinsic generative capacity by evaluating it post-hoc on a fixed set of diverse, held-out offline validation states. This ensures our measurements reflect the true model capacity rather than artifacts of a collapsed online replay buffer. Our empirical dynamics (\autoref{fig:dynamics_actor_critic}, Middle) reveal that as the flow network unlearns the prior, its internal representations rapidly degenerate, causing the rank of its features to plummet. This representational collapse directly destroys the policy's generative capacity, leading to an immediate and sharp drop in both cross-state and within-state action diversity. While cross-state diversity captures whether the policy adapts to different states, within-state diversity ensures the generative model can produce multiple valid actions for a single state. This broad distribution preserves local exploration and, particularly for IFQL, provides the critic with a diverse menu of candidate actions to evaluate. Stripped of this exploratory power, the agent's behavior undergoes a severe spatial collapse. PCA projections of online rollouts confirm this restriction, showing the actor trapped in repetitive failure modes with severely limited state-space coverage (\autoref{fig:state_coverage}, Left). A concurrent drop in temporal coherency ($\mathbb{E}[\Vert{}x_{t+5} - x_t\Vert{}]$) for \texttt{cube-double} physically captures this collapse, demonstrating that broad exploration degenerates into localized jitter (\autoref{fig:state_coverage}, Right).

\paragraph{Data starvation breaks the critic.}
This severely restricted coverage explains the downstream failure of value learning (\autoref{fig:dynamics_actor_critic}, Right). The BC objective does not directly alter the critic’s loss landscape, and the only mechanism through which BC can impact the critic is through the data the actor collects. Because the actor's spatial exploration has collapsed, it feeds the critic a narrow, repetitive diet of transitions. Starved of diverse data, the critic's own internal representations degenerate, leading to a severe collapse in its feature rank. We believe this representational collapse directly causes the decreasing critic separation gap, the difference in Q-values assigned to state-actions from top-20\% versus bottom-20\% return trajectories. The critic simply loses its ability to distinguish good states from bad ones because it no longer sees a sufficiently diverse state-action manifold to maintain rich, discriminative features. \textsc{BC-Off} prevents this initial diversity collapse, supplying the critic with the rich, stable data necessary to preserve its feature rank and drive continuous policy improvement.

\begin{figure}[htbp]
    \centering
    \includegraphics[width=0.325\linewidth]{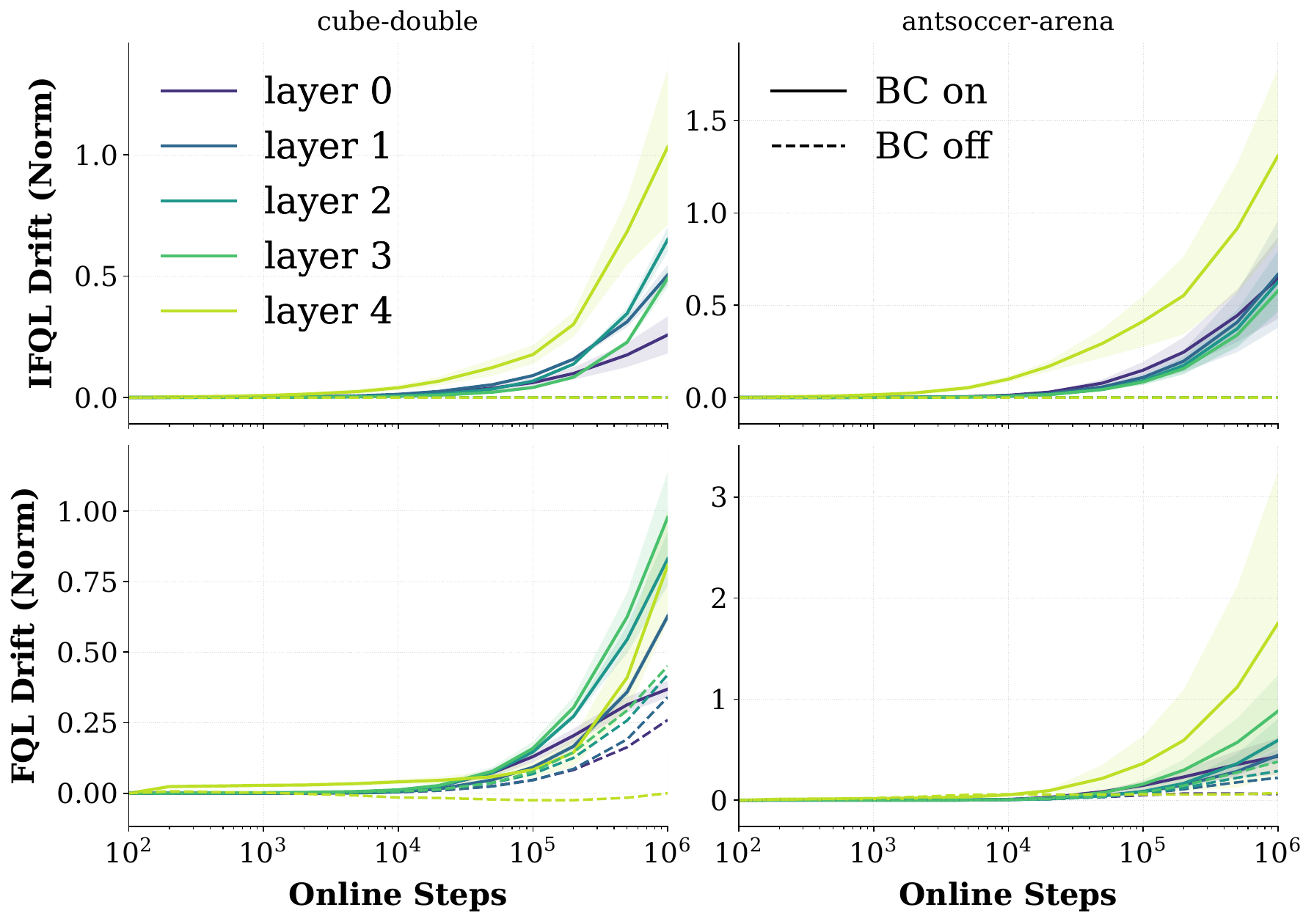}
    \hfill
    \includegraphics[width=0.325\linewidth]{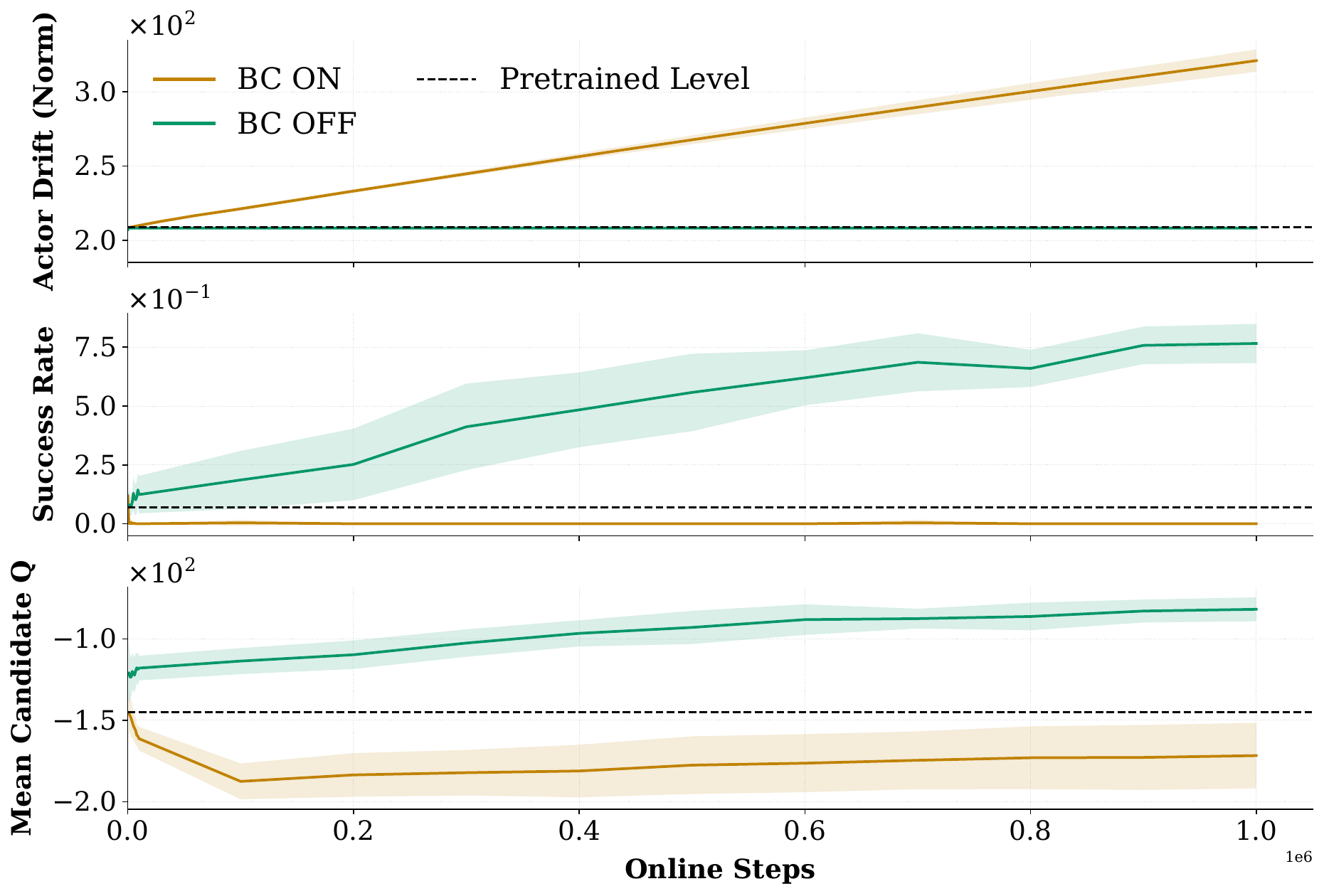}
    \includegraphics[width=0.325\linewidth]{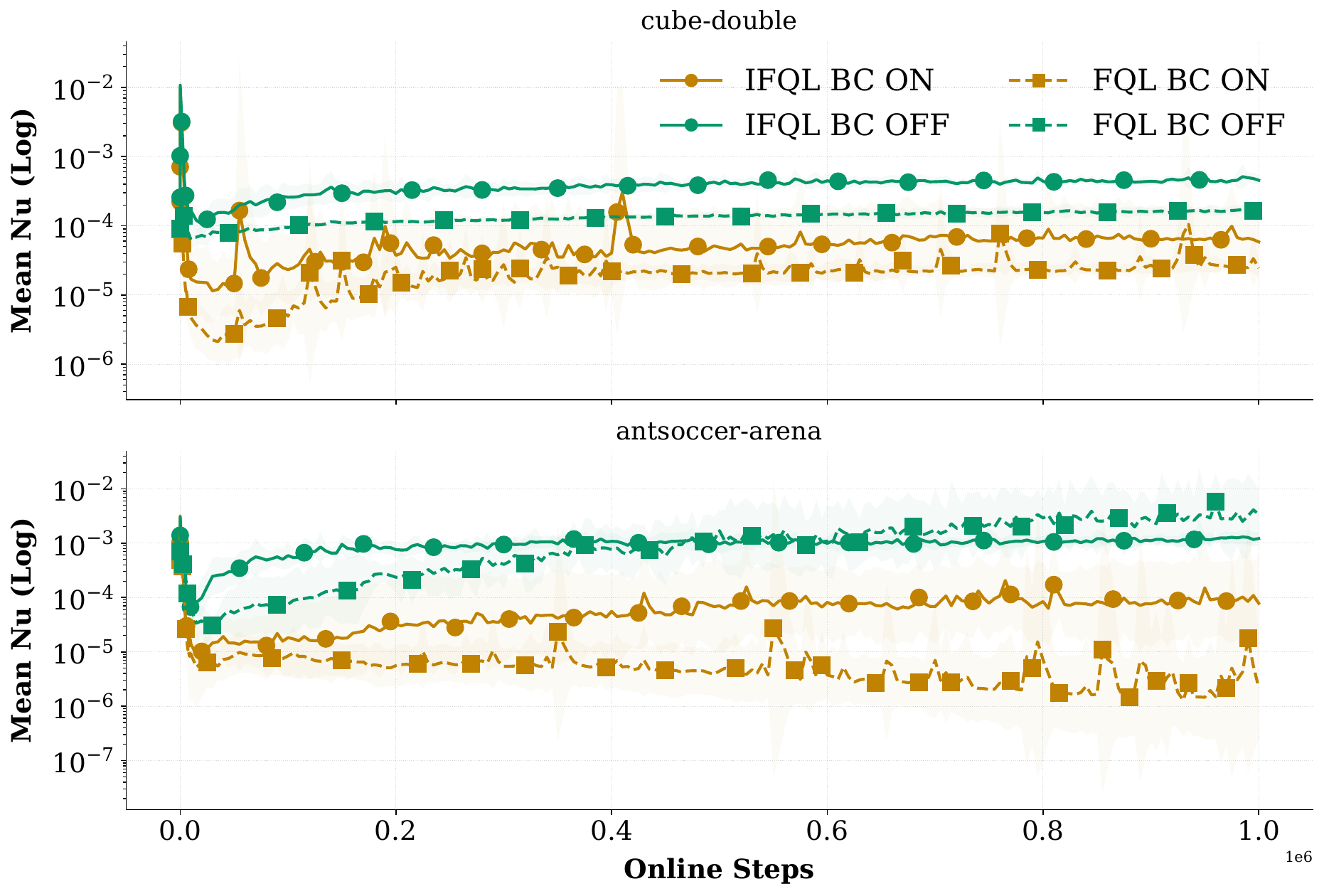}
    \vspace{-0.4cm}
    \caption{
    \textbf{Online BC drives progressive parameter drift and candidate collapse.}
    \textbf{Left:} Per-layer relative drift (log $x$-axis). Under \textsc{BC-On} (solid), drift penetrates chronologically, beginning with the output-adjacent layer ($10^3$--$10^4$ steps). \textsc{BC-Off} (dashed) exactly freezes IFQL and anchors FQL via distillation.
    \textbf{Middle:} Macroscopic drift and downstream impact for IFQL on \texttt{cube-double}. Under \textsc{BC-On}, unbounded parameter drift occurs from the pretrained initialization. \textbf{Right:} Adam's second-moment estimate ($\hat{v}$, log scale). Under \textsc{BC-On} (solid), gradient magnitude plummets immediately at the onset of online training and remains orders of magnitude below \textsc{BC-Off} (dashed), indicating a weakened optimization signal.
    }
    \label{fig:parameter_drift_and_collapse}
\end{figure}

\paragraph{Drift accumulation and non-linear prior degradation.}
Table~\ref{tab:bc_stopping_times} shows that intermediate BC termination yields substantial, non-monotonic recovery. Performance drops sharply from the immediate-freeze optimum ($\tau=0$), plateaus for hundreds of thousands of steps, and collapses under persistent BC ($\tau=1\text{M}$). This trajectory maps directly to the observed parameter drift in (\autoref{fig:parameter_drift_and_collapse}, Left and Middle). The actor's output-adjacent layer diverges within $10^3$--$10^4$ steps, rapidly degrading action quality and permanently lowering the performance ceiling. Deeper representational layers drift more slowly, creating a prolonged window where the prior is damaged but retains sufficient generative capacity to be rescued by removing BC. Persistent BC eventually overwrites these core representations entirely, triggering catastrophic failure.

\paragraph{Ruling out stale-optimizer statistics.}
The failure of online BC is visible at the level of gradient statistics long before downstream metrics degrade. Under no-retention \textsc{BC-On}, Adam's second-moment estimate $\hat{v}$ consistently plummets compared to \textsc{BC-Off} across algorithms and tasks (\autoref{fig:parameter_drift_and_collapse}). Because $\hat{v}$ tracks the running magnitude of squared gradients, this drop indicates that the network is receiving weak, homogeneous gradient signals. Furthermore, because standard offline-to-online RL implementations routinely carry over optimizer statistics from pretraining into the online phase, this sharp divergence introduces a natural alternative hypothesis: the observed collapse might stem from stale Adam moment estimates failing to adapt to the online distribution shift, rather than the BC objective itself. To isolate this confounder, we clear the optimizer state at the transition boundary. Across all six \textsl{Discard}+\textsc{BC-On} configurations, this reset yields exactly zero performance recovery, with final success rates remaining rigidly at 0.000 (\autoref{fig:reset-does-not-explain-collapse}, Left). Under the functional \textsc{BC-Off} regime, the reset produces only marginal and inconsistent fluctuations (\autoref{fig:reset-does-not-explain-collapse}, Right). These results systematically rule out optimization artifacts as a failure mechanism.

\section{Related Work}
\label{sec:related_work}

\paragraph{Behavior regularization in offline-to-online RL.}
Offline RL commonly regularizes policies toward dataset-supported behavior to
limit extrapolation error
\citep{fujimoto2019off,kumar2019stabilizing,levine2020offline}. Methods such as
BCQ, BEAR, BRAC, advantage-weighted regression, TD3+BC, and ReBRAC differ in how
they impose this constraint, but all rely on the dataset to define which
behaviors should be preserved
\citep{fujimoto2019off,kumar2019stabilizing,wu2019behavior,
nair2020awac,kostrikov2021offline,fujimoto2021minimalist,tarasov2023revisiting}. Most offline-to-online methods retain the offline dataset during fine-tuning,
whereas considerably less work studies the setting in which that supervision
disappears. WSRL discards the dataset but stabilizes the transition through
pretrained-policy rollouts, while PORL considers a policy-only initialization
with a newly learned critic
\citep{zhou2025efficient,xiao2025efficient}. We retain the offline-pretrained actor and critic and study behavior
regularization after its defining data become unavailable. Our results show
that behavioral fitting should stop once that supervision disappears. See
App.~\ref{app:related_work} for a broader discussion.

\paragraph{Expressive policies for data-driven RL.}
Diffusion-QL combines an expressive diffusion actor with denoising-based BC
and value maximization \citep{wang2022diffusion}. IDQL instead models the
behavior distribution with diffusion and obtains an improved decision rule by
critic-guided reweighting or selection \citep{hansenestruch2023idql}. FQL uses
conditional flow matching to learn a behavior model and distills value-improved
actions into an efficient one-step actor \citep{park2025flow}. Q-chunking
extends expressive behavior modeling to temporally extended action sequences
\citep{li2025actionchunking}. Related work on policy-agnostic improvement also
separates value improvement from the policy class used to represent the
resulting actions \citep{mark2024policy}. Recent methods such as EXPO and FINO
further investigate stable online optimization with expressive policies
\citep{dong2026expo,shin2026flow}. We study a distinct question: whether an
expressive behavioral model can remain useful as a fixed prior after both its
training data and its behavioral-fitting objective are removed.

\section{Lessons and Opportunities}
\label{sec:discussion}

\textbf{Lessons.}
No-retention fine-tuning is fundamentally distinct from conventional offline-to-online RL. When the offline dataset is discarded, behavioral cloning ceases to act as a stabilizing regularizer; instead, it forces the network to imitate its own narrowing online experience. As shown, this rapidly cannibalizes the behavioral prior, causing a representational collapse that starves the rest of the learning process. The solution is not simply to turn off the objective, but to embrace a two-part design principle: successfully transitioning to online-only data requires (1) freezing an inherently expressive, multimodal behavioral prior, and (2) maintaining a decoupled decision mechanism (such as critic-guided selection or distillation) that allows online value learning to exploit that frozen manifold.

\textbf{Opportunities.}
Our findings open several new avenues for data-restricted RL. While a strictly frozen prior is highly effective when pretraining and fine-tuning share the same underlying dynamics, extreme environmental distribution shifts may eventually require the prior to adapt. This motivates the need for mechanisms that bridge the gap between catastrophic online-only BC and hard freezing. Future work could explore generative replay, uncertainty-aware plasticity, or selective behavioral updates that preserve the broad, valid modes of the offline dataset while safely incorporating novel online behaviors. This is especially relevant to VLA fine-tuning, where expressive pretrained policies may need to support online improvement after demonstration data are discarded \citep{dong2026expoftsampleefficientreinforcementlearning,dong2026reinforcementlearningrealtimevisionlanguageaction}. Ultimately, treating the pretrained model as a permanent, expressive foundation rather than a strictly plastic network offers a robust path forward for RL deployments with strict data constraints.

 \section*{Acknowledgments}
The authors would like to thank Walter Mayor for valuable discussions during the preparation of this work. We also thank Seohong Park for his helpful responses and assistance in clarifying questions regarding the methods and datasets. The research was enabled in part by computational resources provided by the Digital Research
Alliance of Canada (\url{https://alliancecan.ca}) and Mila (\url{https://mila.quebec}). Mohammad Sami Nur Islam was supported by a Master's research scholarship from the Fonds de recherche du Québec – Nature et technologies (FRQNT) under award number 369318. Aaron Courville acknowledges funding from the Institute of Information \& Communications Technology Planning \& Evaluation (IITP) through a grant funded by the Korean government (MSIT) (No. RS-2024-00457882, National AI Research Lab Project). Pablo Samuel Castro acknowledges funding from NSERC Discovery Grant. We want to acknowledge funding support from Google and CIFAR AI. We would also like to thank the Python community \citep{van1995python, 4160250} for developing tools that enabled this work, including NumPy \citep{harris2020array}, Matplotlib \citep{hunter2007matplotlib}, Jupyter \citep{2016ppap}, and Pandas \citep{McKinney2013Python}.

\section*{AI use statement}

In this work, we used generative AI tools for aiding and polishing paragraph-level writing, discussing and refining the structure and flow of the introduction, finding related work, and editing and writing the code. We also used generative AI tools to assist with early drafts of some sections, which were later substantially rewritten and replaced by the authors.
We have not used generative AI tools for research ideation, and the rest of the required disclosure tasks, such as generating synthetic datasets or proving mathematical claims, are not applicable to this work.
We have reviewed the AI-assisted work. We reviewed LLM-generated code and read and incorporated suggested related work papers manually.
We take responsibility for the final content of this work, including text, claims or artifacts produced with the aid of generative AI.

\bibliography{iclr2027_conference}
\bibliographystyle{iclr2027_conference}

\clearpage
\newpage

\appendix
\onecolumn

\etocdepthtag.toc{appendix}
\etocsettocstyle{

\section*{Appendix Contents}}{} % Title of the TOC
\setcounter{tocdepth}{3}
\etocsettagdepth{mtoc}{none}           % Hide main text sections
\etocsettagdepth{appendix}{subsection}  % Show appendix 
\tableofcontents                        % Print the TOC

\clearpage
\newpage

\section{Additional Related Work}
\label{app:related_work}

Offline reinforcement learning is vulnerable to extrapolation error because
value functions and policies may be evaluated on actions that are poorly
represented in the dataset
\citep{fujimoto2019off,kumar2019stabilizing,levine2020offline}.
A broad family of methods therefore constrains policy improvement toward the
offline behavior distribution. BCQ restricts action selection using a learned
behavior model \citep{fujimoto2019off}; BEAR and BRAC penalize divergence from
the behavior policy \citep{kumar2019stabilizing,wu2019behavior}; and
advantage-weighted methods regress toward dataset actions according to their
estimated value
\citep{peng2021advantageweighted,nair2020awac,kostrikov2021offline}.
TD3+BC and ReBRAC combine value maximization with direct regression toward
actions in the replay buffer
\citep{fujimoto2021minimalist,tarasov2023revisiting}. Although these methods
differ in form, they share an important assumption: the data defining the
behavioral constraint remain available during optimization. The behavioral
objective is meaningful because its supervision continues to represent the
desired action distribution. We study what happens when the objective remains
active but the data that originally defined it do not.

Offline-to-online reinforcement learning initializes an online learner from
offline pretraining and subsequently improves it through environment
interaction \citep{li2026the}. AWAC uses offline data to initialize a policy and then performs
advantage-weighted online updates \citep{nair2020awac}. Other methods retain
offline transitions through balanced replay, large update-to-data ratios, or
conservative value estimation
\citep{lee2022offline,ball2023efficient,nakamoto2023calql}. Cal-QL explicitly
calibrates the offline critic so that its values provide a more suitable
initialization for online fine-tuning
\citep{nakamoto2023calql}. Additional methods manage the stability--plasticity
trade-off using separate offline and online policies, adaptive behavioral
constraints, or evolving reference policies
\citep{zhang2023policy,zhao2022adaptive,li2023proto}. These approaches show
that offline supervision can accelerate online learning, but they generally
assume continued access to the offline dataset or introduce an additional
mechanism for preserving its influence.

A smaller body of work studies online adaptation after offline data are
discarded. WSRL shows that offline data need not be retained when the online
replay buffer is initialized with rollouts from the pretrained policy
\citep{zhou2025efficient}. Its analysis emphasizes critic recalibration and
value divergence under the distribution shift from offline data to online
experience. PORL considers an even more restrictive policy-only setting in
which the pretrained policy is available but the offline critic is not, and a
new critic must be learned from interaction
\citep{xiao2025efficient}. Reincarnating RL studies the broader problem of
reusing previously learned policies and value functions to accelerate
subsequent reinforcement learning rather than restarting from scratch
\citep{agarwal2022reincarnating}. These works establish that no-retention
fine-tuning is possible, but primarily focus on preserving or reconstructing
value estimates. We retain the offline-pretrained actor and critic and study a
complementary actor-side question: should behavioral fitting continue once the
data that defined its objective become unavailable?

Another line of work treats offline-to-online fine-tuning as a replay
composition problem. Balanced replay and related methods control the relative
contribution of offline and online transitions through sampling ratios,
prioritization, or adaptive buffer construction
\citep{lee2022offline,ball2023efficient}. These methods primarily change the
data seen by the critic and actor. Our intervention is different: we change
whether the behavioral model continues to optimize an objective whose
supervision has changed. This distinction is important in the no-retention
setting. Once the offline buffer is unavailable, replay manipulation alone
cannot restore the original behavioral distribution unless that distribution
has been preserved by the model itself.

Diffusion- and flow-based policies provide expressive alternatives to
unimodal Gaussian actors and can represent multimodal conditional action
distributions
\citep{wang2022diffusion,liu2022flow,lipman2023flow}. Diffusion-QL combines
behavior modeling with value maximization
\citep{wang2022diffusion}, while IDQL uses a diffusion behavior model to
propose actions and a critic to reweight or select among those proposals
\citep{hansenestruch2023idql}. FQL replaces iterative diffusion with flow
matching and distills value-improved actions into an efficient one-step actor
\citep{park2025flow}. These methods separate the distribution used to propose
actions from the mechanism used to improve or select them.

Recent work also studies how expressive policies should be optimized online.
EXPO avoids directly propagating unstable value gradients through a
long-horizon generative policy and instead uses a lightweight edit policy to
modify actions sampled from an expressive base policy
\citep{dong2026expo}. FINO adapts flow policies online through noise injection
and entropy-guided sampling to promote exploration beyond the offline
distribution
\citep{shin2026flow}. These methods continue adapting the expressive policy
while attempting to stabilize its optimization. We study a different
intervention: preserving the pretrained expressive prior and continuing online
improvement through the critic and decision rule.

Action chunking extends expressive behavior modeling from individual actions to
temporally extended action sequences. Q-chunking applies this idea directly to
reinforcement learning, using chunked actions to preserve temporally coherent
behavior, improve exploration, and enable multi-step value backups
\citep{li2025actionchunking}. This setting is especially relevant to
no-retention fine-tuning because updating a chunk prior on a small online
buffer can erase not only individual actions but also longer-horizon temporal
structure. We therefore evaluate whether the prior-preservation principle
extends from single-action flow policies to action-chunked RL.

Our work differs from these studies in its central intervention and
diagnostic question. We do not introduce a new expressive policy class, a new
critic-calibration procedure, or a new replay rule. Instead, we isolate the
consequences of continuing versus stopping behavioral-model fitting after the
offline data disappear. The resulting comparison reveals when an expressive
prior remains useful, when removing BC is insufficient, and why the connection
between the prior and the deployed decision rule is essential.

\section{Experiment Details}
\label{app:experiment_details}

\subsection{Computational Resources}
\label{app:compute}

To optimize computational efficiency and ensure strictly matched starting conditions, we decoupled the offline pretraining phase from the online fine-tuning phase. For each algorithm, task, and random seed, we first executed the $1$M offline gradient steps and saved the resulting pretrained checkpoints. All subsequent online fine-tuning setups were then initialized from these shared checkpoints. 

Experiments were executed on NVIDIA H100 GPUs. To maximize hardware utilization, we multiplexed six individual experimental runs per GPU, with the underlying allocation providing 12 CPU cores and 124\,GB of system memory under a 12-hour wall-clock budget. In total, the full study includes several hundred individual online training runs across the four evaluated algorithms (FQL, IFQL, ReBRAC, and QC).

Reusing pretrained checkpoints avoids redundant offline computation across fine-tuning configurations, while freezing the behavioral prior eliminates the need for continued behavioral-model optimization. These choices simplify controlled experimentation and may reduce computational overhead, supporting broader efforts toward accessible and reproducible deep RL research \citep{ceron2021revisiting}. Such efficiency considerations are particularly relevant when conducting systematic investigations of architectural choices \citep{willi2024mixture,pmlr-v235-obando-ceron24a,pmlr-v235-obando-ceron24b,obando2026simplicial} and experimental configurations \citep{obando2023small,schwarzer2023bigger}.

\subsection{Implementation and Hyperparameters}
\label{app:hyper}

We base our implementations of FQL, IFQL, and ReBRAC on the official FQL repository, and Action-Chunked RL (QC) on the official QC repository. To ensure fair evaluations, we do not re-tune the base optimization hyperparameters. Instead, we retain the standard architecture and optimization settings from the reference implementations and apply their recommended task-specific hyperparameters. These values remain strictly fixed across all experiments so that our ablations isolate only the specific architectural or algorithmic components under study. Nevertheless, deep RL
agents can be sensitive to hyperparameter selection
\citep{ceron2024on}, and independently tuning each
configuration may yield different results at substantial
computational expense.

For FQL, the actor, flow model, and critic use four-layer MLPs with hidden dimensions $(512,512,512,512)$ and GELU activations. Optimization uses Adam with a learning rate of $3\times10^{-4}$, minibatches of size $256$, and a target-network smoothing coefficient $\tau=0.005$. Flow policies use linear flow-matching paths, uniformly sampled flow times, and $10$ Euler integration steps. FQL uses two Q-functions; their mean is used in the actor objective and, for the OGBench tasks considered here, also for the critic target.

IFQL uses the same flow-matching configuration as FQL and follows the default IQL value-learning setup, utilizing an IQL expectile of $\kappa = 0.9$. At evaluation time, IFQL samples $N$ candidate actions from the behavioral flow prior and executes the action with the highest estimated value. We use the task-specific values of $N$ detailed in Table~\ref{tab:task_hparams}.

For QC, we preserve the flow-matching setup and maintain a fixed action chunking horizon of $H = 5$ across all tasks. Similar to IFQL, QC evaluates by sampling $N$ candidate action sequences from the frozen prior and selects the sequence that maximizes the chunk-level critic.

ReBRAC uses identical default architecture and optimization settings unless otherwise specified. Its primary method-specific hyperparameters are the actor behavioral-cloning coefficient $\alpha_1$ and the critic behavioral-cloning coefficient $\alpha_2$.

\begin{table}[htbp]
\centering
\small
\setlength{\tabcolsep}{4pt}
\renewcommand{\arraystretch}{1.12}
\caption{Base architecture and optimization hyperparameters.}
\label{tab:shared_hparams}
\begin{tabular}{@{}ll@{}}
\toprule
\textbf{Hyperparameter} & \textbf{Value} \\
\midrule
Default MLP hidden dimensions & $(512,512,512,512)$ \\
Activation & GELU \\
Optimizer & Adam \\
Learning rate & $3\times10^{-4}$ \\
Batch size & $256$ \\
Target-network update rate $\tau$ & $0.005$ \\
Discount $\gamma$ & $0.99$ default; $0.995$ for humanoidmaze/antsoccer \\
Number of Q-functions & $2$ \\
\midrule
\multicolumn{2}{@{}l}{\textit{FQL / IFQL / QC flow policy}} \\
Flow time sampling distribution & $\mathrm{Unif}([0,1])$ \\
Euler integration steps & $10$ \\
\midrule
\multicolumn{2}{@{}l}{\textit{IFQL}} \\
IQL expectile $\kappa$ & $0.9$ \\
\midrule
\multicolumn{2}{@{}l}{\textit{QC}} \\
Action chunking horizon $H$ & $5$ \\
\bottomrule
\end{tabular}
\end{table}

\begin{table}[htbp]
\centering
\small
\setlength{\tabcolsep}{4pt}
\renewcommand{\arraystretch}{1.12}
\caption{Task-specific hyperparameters adopted from the FQL and QC
papers. IFQL and QC report the number of candidate actions or sequences $N$,
FQL reports its behavioral-regularization coefficient $\alpha$, and ReBRAC reports
the actor and critic BC coefficients $(\alpha_1,\alpha_2)$.}
\label{tab:task_hparams}
\begin{tabular}{@{}lcccc@{}}
\toprule
\textbf{Task}
& \textbf{IFQL $N$}
& \textbf{FQL $\alpha$}
& \textbf{QC $N$}
& \textbf{ReBRAC $(\alpha_1,\alpha_2)$} \\
\midrule
\texttt{cube-double}
& $32$
& $300$
& $32$
& $(0.1, 0)$ \\

\texttt{puzzle-4x4}
& $32$
& $1000$
& $32$
& $(0.3, 0.01)$ \\

\texttt{antsoccer-arena}
& $64$
& $30$
& $64$
& $(0.01, 0.01)$ \\

\texttt{humanoidmaze-medium}
& $32$
& $100$
& $32$
& $(0.01, 0.01)$ \\

\texttt{humanoidmaze-large}
& $32$
& $100$
& $32$
& $(0.01, 0.01)$ \\

\bottomrule
\end{tabular}
\end{table}

\begin{table}[htbp]
\centering
\small
\setlength{\tabcolsep}{4pt}
\renewcommand{\arraystretch}{1.15}
\caption{Training and evaluation protocol used in our experiments.}
\label{tab:training_protocol}
\begin{tabular}{@{}p{0.38\columnwidth}p{0.56\columnwidth}@{}}
\toprule
\textbf{Setting} & \textbf{Value} \\
\midrule
Offline pretraining steps
& $1{,}000{,}000$ \\

Online fine-tuning steps
& $1{,}000{,}000$ \\

Replay buffer capacity
& $2{,}000{,}000$ transitions \\

Evaluation episodes per checkpoint
& $50$ \\

Evaluation interval
& Every $100{,}000$ steps \\

Seeds per configuration
& 3 \\
\bottomrule
\end{tabular}
\end{table}

\newpage

\subsection{Domain Details}
\label{app:domain}
\subsubsection{OGBench}
We cover both manipulation and locomotion domains from the \textsc{OGBench} benchmark using the single-task variants of each environment. We use all five tasks available for each domain. Under the single-task variant, the reward structure is semi-sparse: locomotion tasks receive a binary 0 or 1 reward upon completion, while manipulation tasks receive a negative reward corresponding to the number of unaccomplished subtasks.

\begin{itemize}
    \item {\texttt{cube-double-play}:} A manipulation task to control a robot arm to arrange two cubes into designated configurations within a 500-step episode.
    \item {\texttt{puzzle-4x4-play}:} A manipulation task requiring the agent to press physical buttons to solve a 4$\times$4 ``Lights Out'' puzzle within a 500-step episode.
    \item {\texttt{scene-play}:} A manipulation task requiring a robot arm to manipulate diverse everyday objects, such as a drawer, window, and button locks, into a desired configuration within a 750-step episode.
    \item {\texttt{antsoccer-arena-navigate}:} A locomotion task where an Ant agent must dribble a soccer ball to a target location in an open arena within a 1000-step episode.
    \item {\texttt{antmaze-large-naviate}:} A locomotion task requiring an 8-DoF Ant agent to navigate to a goal location in a large maze layout within a 1000-step episode.
    \item {\texttt{humanoidmaze-medium-navigate}:} A locomotion task requiring a 21-DoF Humanoid agent to navigate to a goal location in a medium-sized maze within a 2000-step episode.
    \item {\texttt{humanoidmaze-large-navigate}:} A locomotion task requiring the 21-DoF Humanoid agent to navigate to a goal location in a larger and more complicated maze layout within a 2000-step episode.
\end{itemize}

\begin{table}[h]
\centering
\caption{
\textbf{FQL requires a connection to the frozen flow.}
Final success under no-retention fine-tuning.
Freezing the behavioral flow succeeds only when the one-step actor remains
distilled toward it.
}
\label{tab:fql_true_freeze}
\begin{tabular}{lccc}
\toprule
Task & \textsc{BC-On} & \textsc{BC-Off} & \textsc{BC-Off}, no distillation \\
\midrule
\texttt{cube-double}
    & $0.000$ & $\mathbf{0.977\pm0.014}$ & $0.000$ \\
\texttt{antsoccer-arena}
    & $0.000$ & $0.303\pm0.206$ & $0.000$ \\
\bottomrule
\end{tabular}
\end{table}

\subsubsection{Robomimic}

\label{app:robomimic}

To evaluate whether the collapse of the behavioral prior generalizes beyond synthetic benchmarks, we evaluate on the Robomimic benchmark \citep{robomimic2021} using human-collected teleoperation datasets. We adopt the Multi-Human (\texttt{MH}) datasets collected via RoboTurk, where demonstrations were gathered across six human operators stratified into three distinct proficiency levels (experienced, adequate, and inexperienced; 50 demonstrations per operator). Each task dataset contains 300 successful demonstrations featuring natural variance, suboptimal paths, and multimodal execution strategies. We evaluate on the following tasks:

\begin{itemize}
    \item \textbf{\texttt{Lift}:} Requires a 7-DoF Franka Emika Panda arm to lift a small cube above a height threshold.
    \item \textbf{\texttt{Can}:} Requires the robot arm to grasp a coke can and place it inside a designated container bin.
\end{itemize}

\subsection{Evaluation Across Additional Domains}
\label{app:more_domains}

\begin{figure}[h]
    \centering
    \includegraphics[width=\textwidth]{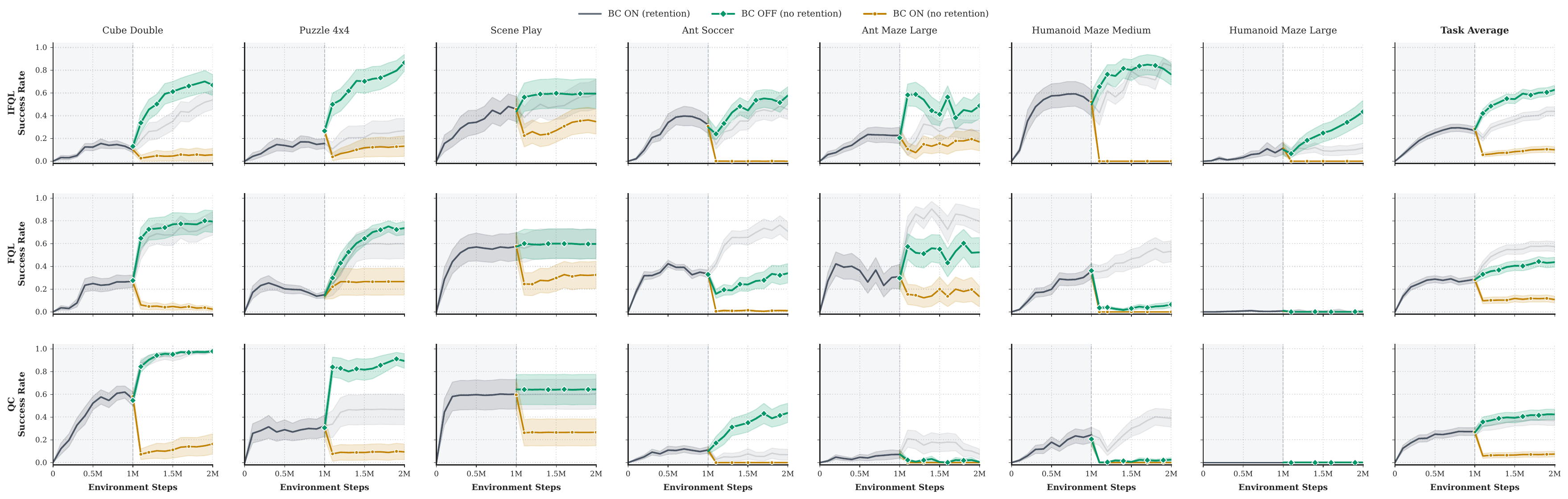}
    \caption{\textbf{Online fine-tuning across all seven OGBench domains.}
    Performance across 35 tasks (5 per domain) for IFQL, FQL, and QC.}
    \label{fig:appendix_more_domains}
\end{figure}

We extend our evaluation to an additional two OGBench domains by incorporating
\texttt{Scene Play} and \texttt{Ant Maze Large} as can be seen in Fig~\ref{fig:appendix_more_domains}.

\section{Mechanistic Analysis Metric Definitions}
\label{app:metrics_definitions}

In \cref{sec:analysis}, we track several metrics to quantify the representational and behavioral collapse of the actor and critic. Prior work has highlighted the importance of
representation quality, effective network capacity, and
stable gradient propagation for deep RL
\citep{obando2026representation,obando2026simplicial,
pmlr-v235-obando-ceron24a,castanyer2025stable,
liu2026measure}. Complementary studies have also shown
that the quality and diversity of collected experience
can substantially affect learning dynamics
\citep{mayor2025the,liu2025the}. To decouple the intrinsic capacity of the networks from the evolving quality of the online replay buffer, these metrics are evaluated post-hoc on a fixed, held-out offline validation set.

\paragraph{Actor Feature Rank.}
This metric measures the effective dimensionality of the actor network's penultimate representation. We pass 1,024 offline validation states through the actor and extract the penultimate layer activations (dimension 512) to form a feature matrix $X$. We center the features ($X_{\text{centered}} = X - \mu_X$) and compute the singular values $s_i$ using SVD. We compute the threshold stable rank \citep{kumar_implicit_2021} as the minimum number of singular values required to capture 99\% of the total singular value energy: $\sum_{i=1}^{k} s_i / \sum_i s_i \geq 0.99$. A higher rank indicates that the policy maintains rich feature representations without collapsing into a low-dimensional subspace.  This metric
is motivated by recent findings linking representation
dimensionality and geometry to learning stability under
non-stationarity \citep{pasand2026stable}.

\paragraph{Action Diversity (Within State).}
This metric measures the proposal stochasticity and sensitivity of the actor to input noise at a given state. For 128 validation states, we sample multiple action candidates per state using different random noise seeds. We compute the standard deviation of the actions at each state and report the mean across all states. High within-state diversity indicates the generative model can produce multiple distinct valid actions; low values indicate a deterministic collapse.

\paragraph{Action Diversity (Between States).}
This metric measures the spatial dispersion of predicted actions across different states. We predict the mean action $a_i = \pi(s_i)$ for $K=128$ distinct validation states and compute the mean pairwise Euclidean distance across all state pairs: $\frac{1}{K(K-1)} \sum_{i \neq j} \|a_i - a_j\|_2$. If this metric collapses near zero, the actor is outputting a constant action regardless of the input state.

\paragraph{Critic Feature Rank.}
This metric measures the effective representation rank of the Q-function network. We pass 1,024 offline $(s, a)$ pairs through the critic, extract the penultimate layer activations, and compute the threshold stable rank exactly as done for the actor (with a 99\% cumulative energy threshold). This measures whether the critic maintains expressive, high-capacity features capable of accurate value prediction. Prior work has demonstrated the importance of structured
representations for value learning and actor-critic
sample efficiency
\citep{obando2026representation,obando2026simplicial},
while studies of network capacity and optimization
highlight additional factors governing representation
quality and adaptability
\citep{pmlr-v235-obando-ceron24a,castanyer2025stable,
liu2026measure}.

\paragraph{Critic Separation Gap}
This metric quantifies the ability of the critic to separate high-quality transitions from poor ones. We partition the offline validation dataset by trajectory return, isolating the top 20\% of transitions ($(s_{\text{top}}, a_{\text{top}})$) and the bottom 20\% ($(s_{\text{bot}}, a_{\text{bot}})$). We evaluate the ensemble minimum Q-values for both subsets and compute the difference: $\mathbb{E}[Q(s_{\text{top}}, a_{\text{top}})] - \mathbb{E}[Q(s_{\text{bot}}, a_{\text{bot}})]$. A large positive gap indicates the critic correctly assigns higher values to high-return states, whereas a gap collapsing to zero indicates severe value distortion.

\section{Experiments}
\label{app:experiments}

\paragraph{Optimizer-state reset.}
One possible explanation for the offline-to-online collapse is that the
optimizer carries stale Adam statistics from offline pretraining into a
different online data distribution. If this were the primary cause, resetting
Adam's first- and second-moment estimates at the offline-to-online transition
should restore performance, particularly in the failing
\textsc{Discard}+\textsc{BC-On} condition.

\begin{figure}[h]
    \centering
    \includegraphics[width=0.8\linewidth]{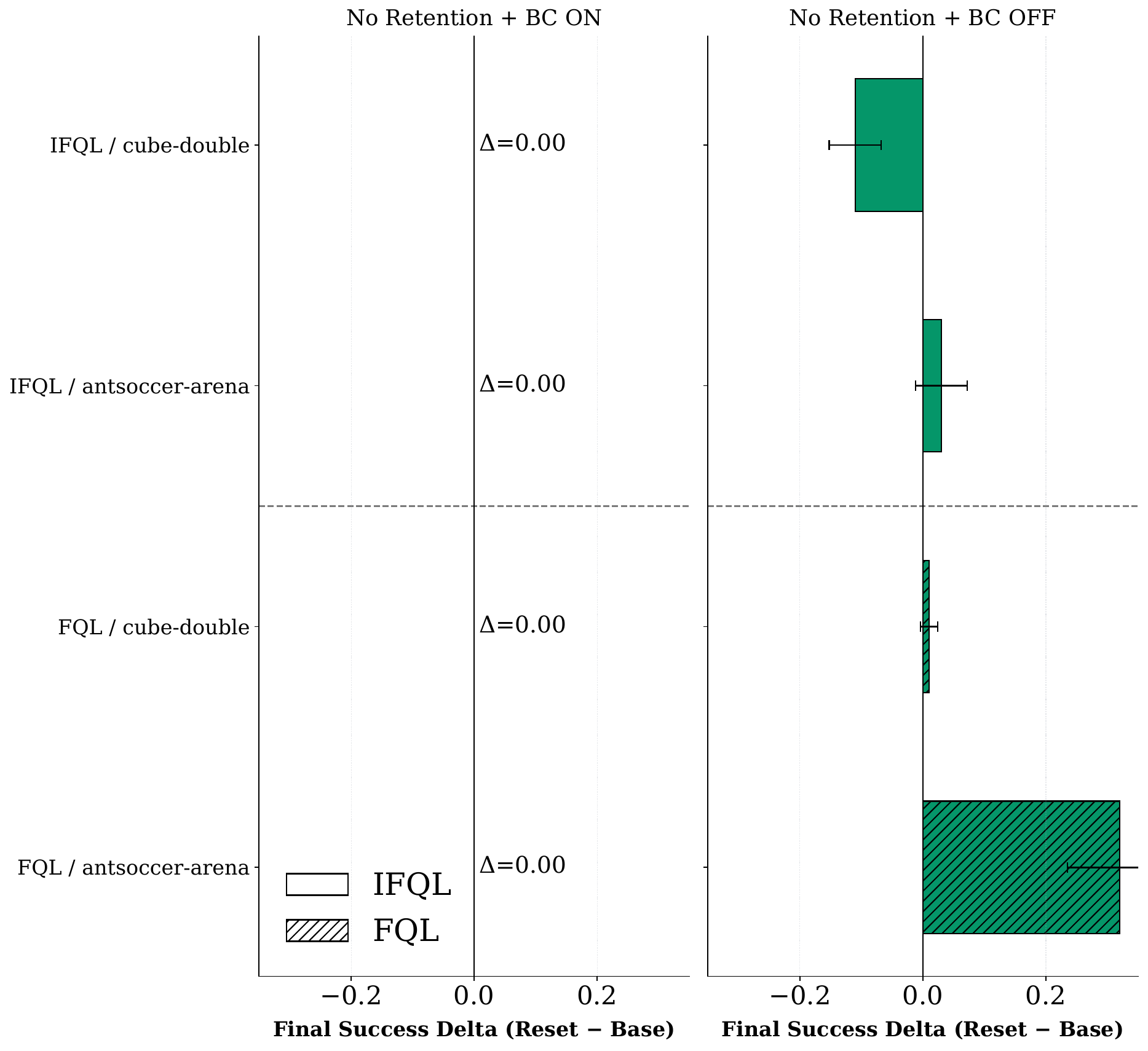}
    \caption{
    \textbf{Resetting the optimizer's Adam state} at the online switch never rescues the discard+BC-on collapse (left, all six cells at exactly zero change) and has only small, inconsistent effects once BC is already off (right). The failure mode is not stale optimizer statistics.}
    \label{fig:reset-does-not-explain-collapse}
\end{figure}

We therefore repeat the corresponding experiments with the optimizer state
reset immediately before online fine-tuning, while keeping the pretrained
network parameters and all other hyperparameters unchanged. As shown in
Figure~\ref{fig:reset-does-not-explain-collapse}, resetting Adam does not
recover performance in any of the six \textsc{Discard}+\textsc{BC-On} cells:
all remain at zero change. Under \textsc{BC-Off}, the reset produces only small
and inconsistent differences. These results indicate that stale optimizer
statistics do not explain the collapse; the dominant failure is associated
with continuing behavioral fitting on online-only data.

\end{document}